\documentclass{article}

\usepackage{PRIMEarxiv}

\usepackage[utf8]{inputenc} % allow utf-8 input
\usepackage[T1]{fontenc}    % use 8-bit T1 fonts
\usepackage{hyperref}       % hyperlinks
\usepackage{url}            % simple URL typesetting
\usepackage{booktabs}       % professional-quality tables
\usepackage{amsfonts}       % blackboard math symbols
\usepackage{nicefrac}       % compact symbols for 1/2, etc.
\usepackage{microtype}      % microtypography
\usepackage{lipsum}
\usepackage{xcolor}
\definecolor{memeYellow}{RGB}{240,242,230}
\usepackage{subcaption}
\usepackage{pgfplots}
\pgfplotsset{compat=1.18}
\usepackage{pgf-pie}
\usepackage{pgfplots}
\pgfplotsset{compat=1.18}
\usepackage{tikz}
\usepackage{tabularx}
\usepackage[table]{xcolor}
\usepackage{amsmath}
\usepackage{fancyhdr}       % header
\usepackage{graphicx}       % graphics
\graphicspath{{media/}}     % organize images and other figures under media/ folder

\title{Multimodal Climate Disinformation Detection: Integrating Vision-Language Models with External Knowledge Sources}

\author{
  Marzieh Adeli Shamsabad\\
  CRIM\\
  \texttt{marzieh.adeli-shamsabad@crim.ca} \\
  \And
  Hamed Ghodrati  \\
  CRIM \\
  \texttt{hamed.ghodrati@crim.ca} \\
  \And
   \\
   \\
  \texttt{} \\
}

\begin{document}
\maketitle

\begin{abstract}
Climate disinformation has become a major challenge in today’s digital world, especially with the rise of misleading images and videos shared widely on social media. These false claims are often convincing and difficult to detect, which can delay actions on climate change. While vision-language models (VLMs) have been used to identify visual disinformation, they rely only on the knowledge available at the time of training. This limits their ability to reason about recent events or updates. The main goal of this paper is to overcome that limitation by combining VLMs with external knowledge. By retrieving up-to-date information such as reverse image results, online fact-checks, and trusted expert content, the system can better assess whether an image and its claim are accurate, misleading, false, or unverifiable. This approach improves the model’s ability to handle real-world climate disinformation and supports efforts to protect public understanding of science in a rapidly changing information landscape.
\end{abstract}

\section{Introduction}

Climate change is one of the most urgent challenges of our time. Although scientific research has built a strong foundation for understanding the crisis, public response is often disrupted by the spread of disinformation. In recent years, this has increasingly taken the form of images and videos that are altered, taken out of context, or even generated by artificial intelligence. When combined with misleading text, such content can make false claims appear convincing and trustworthy, especially on fast-moving platforms like social media.

This new type of disinformation is not just textual, it is visual and multimodal. It spreads quickly, triggers emotional reactions, and is often more difficult to verify. In the context of climate communication, where public trust in evidence is critical, this creates a serious risk.

This paper addresses that challenge by going beyond current vision-language models, which rely solely on internal knowledge acquired during training. These models are unable to reason effectively about recent events or unseen claims. To overcome this limitation, we propose a system that integrates external knowledge sources, allowing the model to access up-to-date information and make more accurate decisions.

The main objectives of this work are:
\begin{enumerate}
    \item Detect climate-related disinformation by analyzing both images and text.
    \item Combine multiple external knowledge sources including reverse image search, web search using Google and GPT, and climate-specific fact-checking websites to verify image–claim pairs.
    \item Use GPT-4o, the latest model with both vision and language understanding, to analyze image–text pairs in context and classify them as accurate, misleading, false, or unverifiable.
\end{enumerate}

The goal is to build a system that improves disinformation detection and supports public understanding of climate science in an online environment where facts are frequently distorted.

\section{Prior Work}\vspace{-1.5ex}

The spread of climate disinformation has become a major concern as people increasingly rely on online platforms for news and scientific information. Researchers have addressed this problem from different angles, developing methods to detect misleading narratives, creating datasets that capture the specific features of climate communication, and testing the capabilities of large language models in this domain.

The study by Coan et al.\ \cite{coan2021} used a hierarchical deep learning model trained on more than eighty seven thousand annotated paragraphs to classify over two hundred forty nine thousand documents into a detailed taxonomy of climate claims. Their analysis revealed patterns that linked certain misinformation themes to funding from conservative donors. Meddeb et al.\ \cite{meddeb2022} focused on the French media landscape, creating a dataset of more than two thousand three hundred climate-related news articles. By combining a CamemBERT model with handcrafted linguistic features, they achieved strong classification results and applied LIME to make the model’s reasoning more transparent. In a broader perspective, Herasimenka et al.\ \cite{herasimenka2023} reviewed thirty eight studies on manipulative climate communication across social media. They found that most work concentrated on platforms such as Twitter and Facebook, rarely tested interventions, and paid little attention to visual platforms like Instagram or TikTok. 

Other researchers have developed specialised detection systems. Rojas et al.\ \cite{rojas2024} introduced Augmented CARDS, which first distinguished between convinced and contrarian tweets before assigning contrarian content to specific categories. Applied to more than five million climate tweets, this approach improved both binary and detailed classification compared to earlier systems. Fore et al.\ \cite{fore2024} examined how large language models can be influenced by false climate claims and evaluated strategies to reduce such effects. They showed that unlearning techniques could substantially lower the rate of contradictions, while retrieval-augmented generation improved truthfulness without modifying model weights. Allaham et al.\ \cite{allaham2025} compared sixteen large language models against expert-labelled data, showing that fine-tuning GPT 3.5 turbo led to higher agreement with experts than both GPT 4o and a BERT-based classifier.

Beyond work that focuses specifically on climate content, a growing body of research in multimodal misinformation detection has explored how external evidence can improve factual verification. LEMMA by Xuan et al.\ \cite{lEMMA} demonstrated how combining the internal reasoning of GPT 4V with documents retrieved through web search and image provenance from reverse image search can enhance classification accuracy. The CMIE framework \cite{CMIE} tackled the challenge of out-of-context misinformation by retrieving contextual evidence from the web and image titles, identifying semantic links between image–text pairs, and weighting the most relevant evidence. Shen et al.\ \cite{shen2025gamed} developed GAMED, which refined multimodal features through expert networks, balanced the importance of each modality, and incorporated semantic knowledge from ERNIE2.0 to achieve high accuracy on benchmark datasets. Wu et al.\ \cite{wu2025} proposed DECEPTIONDECODED, a benchmark where each claim is paired with a trusted reference article. Their results showed that even when evidence is available, leading models often fail to use it effectively to identify misleading intent.

Together, these studies show that climate disinformation detection has advanced considerably, yet most work remains focused on text, specific languages, or a limited set of platforms. At the same time, multimodal approaches that integrate external evidence have rarely been applied to climate-related claims, and even when such evidence is available, models often struggle to make full use of it. This gap points to the need for approaches that bring together climate-specific detection with effective external knowledge integration, enabling more accurate and context-aware verification across different media formats.

\section{Approach}

\subsection{Dataset}

The experiments in this work are based on the CliME (Climate Change Multimodal Evaluation) dataset \cite{CliMe}, which contains 2,579 climate-related social media posts collected from Twitter and Reddit. Each post consists of a short textual claim, an accompanying image, and a manually validated description. The claim corresponds to the main text of the post, along with the supportive image. The descriptions, generated using DeepSeek Janus Pro and reviewed by experts, summarize the overall message conveyed by both the claim and the image. The dataset covers a diverse range of multimodal content, including memes, infographics, and posts expressing either skepticism or support toward climate change.

Since the original dataset does not include factuality labels, we automatically extracted labels by jointly considering the claim, the accompanying image, and the expert-validated description. The extraction process was carried out using GPT-4o with four distinct role based prompt designs, each framing the task from a different perspective to encourage robust reasoning over both modalities. The description was used only during label extraction to assist in interpreting the intended meaning of the multimodal post. For evaluation, only the image and the claim were retained, in order to better simulate real-world fact-checking scenarios where such contextual descriptions are typically unavailable.

Label extraction was performed on a subset of 1,200 image–claim pairs in two different setups, 4-class and 2-class schemes. The number of processed samples was determined by cost-efficiency considerations, as each pair was evaluated using multiple large language model prompts. The selected subset was chosen to increase the proportion of underrepresented classes for example (\textit{False} and \textit{Unverifiable}) in 4-class setup and thereby support the construction of a balanced evaluation set. Samples were processed in multiple batches, with each pair assessed using four distinct prompt designs, each reflecting a different evaluative perspective to encourage robust multimodal reasoning:
\begin{itemize}
    \item \textbf{Prompt 1}: Neutral perspective, encouraging balanced assessment and caution toward emotionally persuasive content.
    \item \textbf{Prompt 2}: Climate scientist perspective, emphasizing alignment with authoritative scientific sources such as the IPCC and NASA.
    \item \textbf{Prompt 3}: Climate policy advisor perspective, focusing on whether the image supports or contradicts the claim.
    \item \textbf{Prompt 4}: Fact-checking reviewer perspective, examining potential visual manipulation and contradictions.
\end{itemize}

In 2-class setup one extra ptompt is added which label extraction based on only the expert description. Final labels for 4-class setup were assigned through 3/4 majority voting, requiring agreement from at least three prompts, and mapped to four factuality categories:
\begin{itemize}
    \item \textit{Accurate}: Factually correct and consistent with established scientific consensus.
    \item \textit{Misleading}: Contains elements of truth but omits important context, exaggerates implications, or presents information in a deceptive manner.
    \item \textit{False}: Directly contradicts established facts or scientific evidence.
    \item \textit{Unverifiable}: Too vague, sarcastic, or lacking sufficient detail to be assessed for factuality.
\end{itemize}

\begin{figure}[h]
    \centering
    
    \begin{subfigure}{0.55\textwidth}
        \centering
        \includegraphics[width=\linewidth]{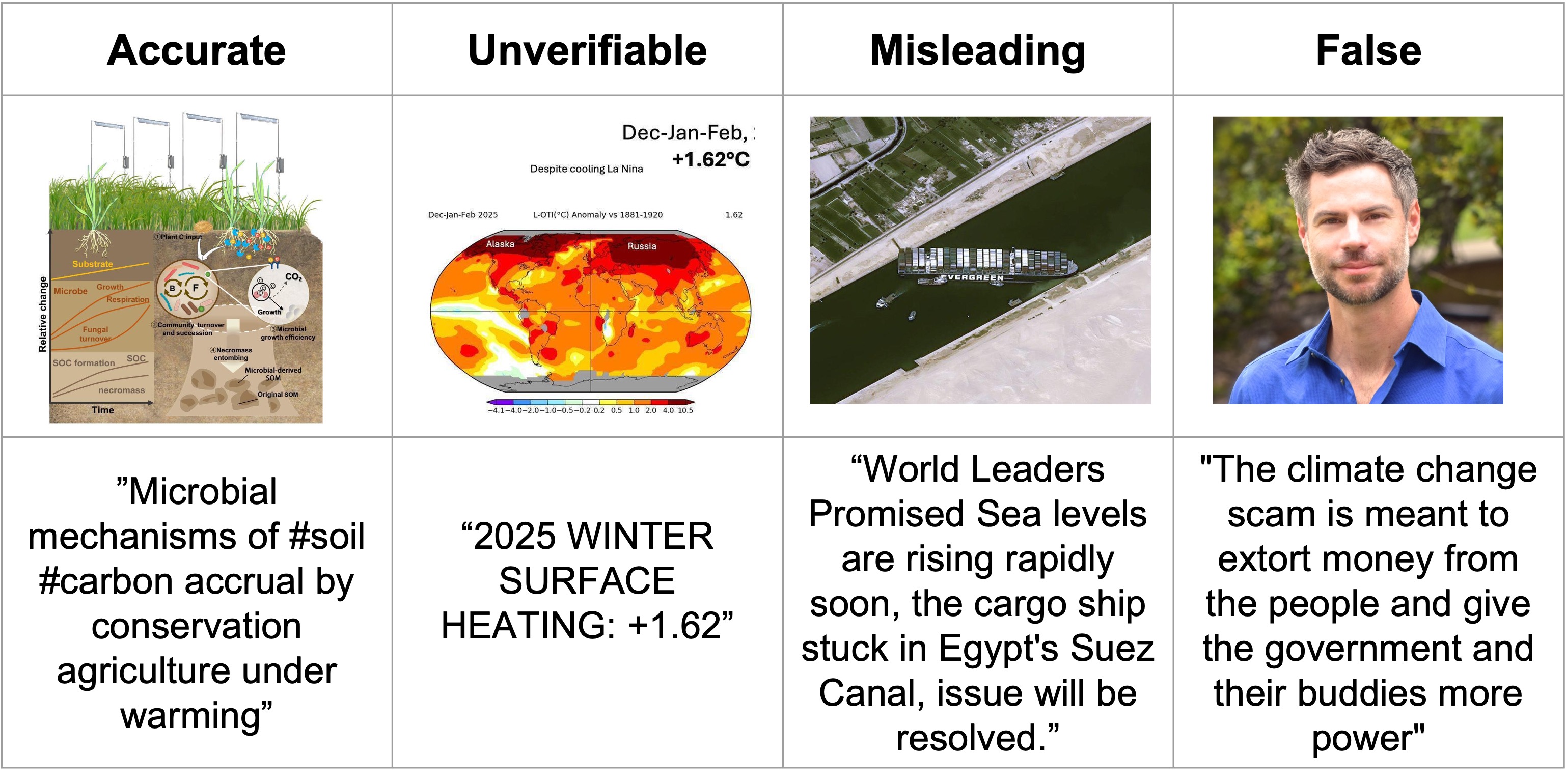}
        \caption{Dataset examples across the 4-class}
        \label{fig:4class_samples}
    \end{subfigure}
    \hfill
    \begin{subfigure}{0.44\textwidth}
        \centering
        \includegraphics[width=\linewidth]{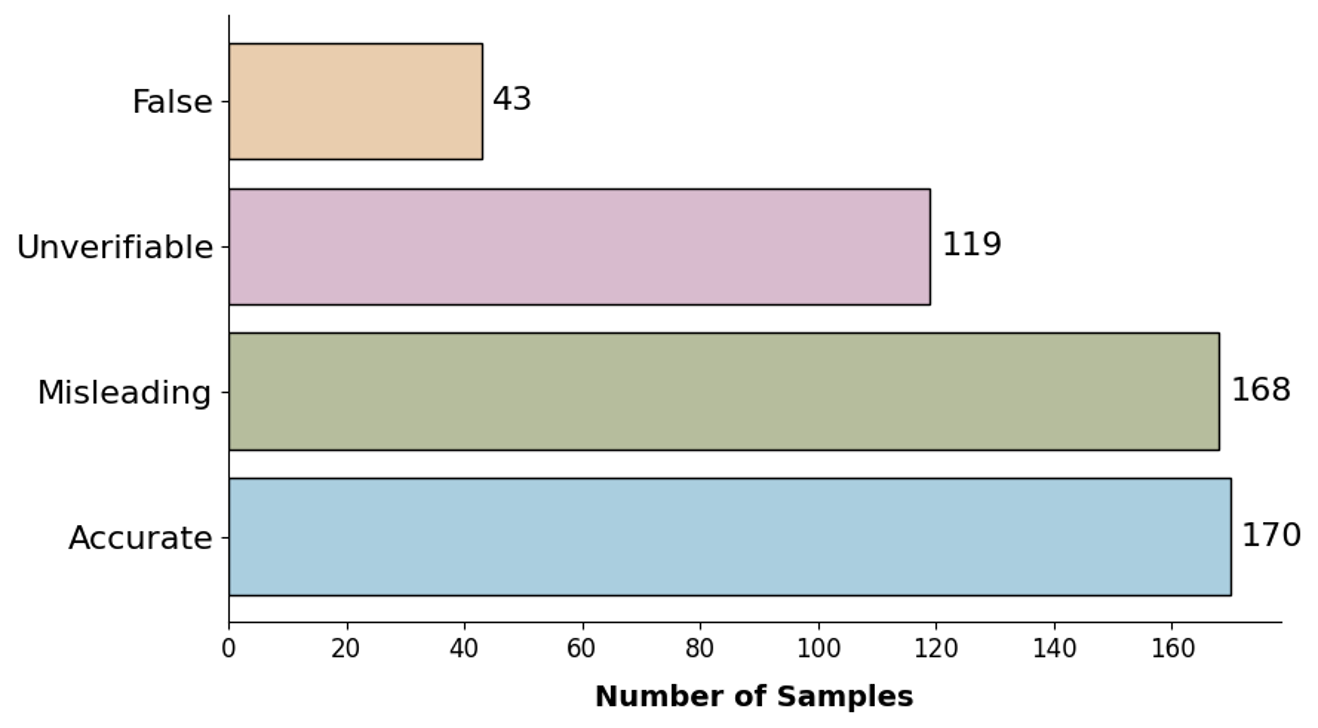}
        \caption{4-class Distribution}
        \label{fig:4class_dis}
    \end{subfigure}
    \caption{Dataset samples and distribution for the 4-class setting}
    \label{fig:4class_dataset}
\end{figure}

\begin{figure}[h]
    \centering
    
    \begin{subfigure}{0.48\textwidth}
        \centering
        \includegraphics[width=\linewidth]{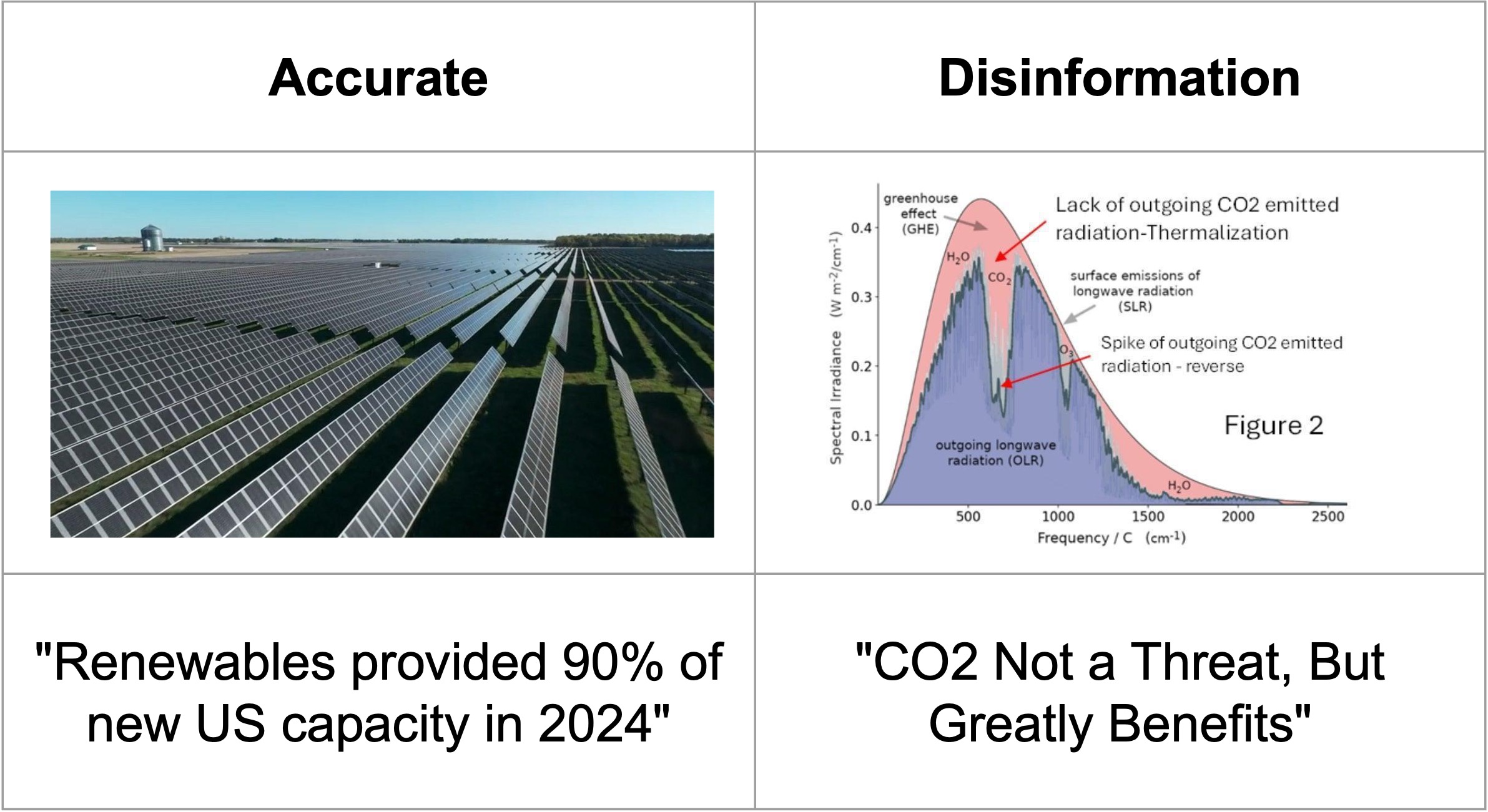}
        \caption{Dataset examples across the 2-class}
        \label{fig:2class_samples}
    \end{subfigure}
    \hfill
    \begin{subfigure}{0.48\textwidth}
        \centering
        \includegraphics[width=\linewidth]{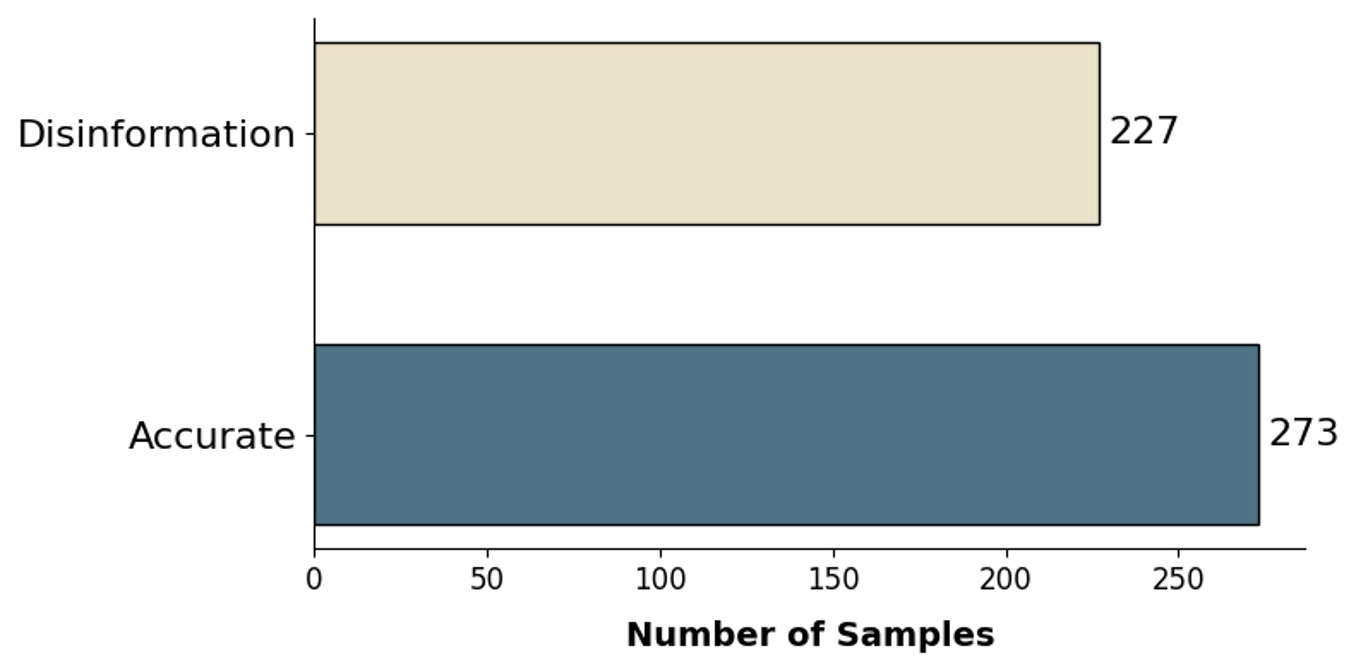}
        \caption{2-class Distribution}
        \label{fig:2class_distri}
    \end{subfigure}
    
    \caption{Dataset samples and distribution for the 2-class setting}
    \label{fig:2class_dataset}
\end{figure}

Final labels for 2-class setup were assigned through 4/5 majority voting, requiring agreement from at least four prompts, and mapped to two factuality categories:
\begin{itemize}
    \item \textit{Accurate}: Factually correct and consistent with established scientific consensus.
    \item \textit{Disinformation}: Contains the samples in unverifiable, misleading or false in this category.
\end{itemize}
The class distribution and the examples of the 4-class dataset are shown in Figure ~\ref{fig:4class_dataset}, and those of the 2-class dataset are shown in Figure ~\ref{fig:2class_dataset}, each has 500  labeled samples.

\subsection{Methodology}

We use GPT-4o as the core model for all experiments. GPT-4o is a multimodal transformer capable of processing both images and text simultaneously, making it well suited for evaluating social media content that combines visual and textual information. Its capacity for image-text reasoning allows it to interpret complex visual elements (e.g., memes, charts, illustrations) in context with accompanying claims.

Our methodology is designed to evaluate the factual consistency between an image and its associated claim. The process is organized into two main stages: input construction with internal knowledge of the VLM and external evidence retrieval.

\begin{itemize}
    \item \textbf{Input Construction:} Each sample in our dataset consists of a claim paired with its corresponding image. This pair forms the core input for evaluation. We prompt GPT-4o to examine the relationship between the visual evidence and the textual claim, with the goal of determining whether the claim is factually supported, misleading, false, or unverifiable.

    \item \textbf{External Evidence Retrieval:} To avoid relying solely on the model’s internal knowledge and to ensure a grounded evaluation, we enhance each sample with evidence gathered from multiple external sources. These sources are complementary, providing both provenance of the image and broader factual context:
    \begin{itemize}
        \item \textbf{Reverse Image Search:} Reverse image search is used to trace the origins of the image and uncover visually similar content published on the web. For each image, we retrieve sources labeled either as exact matches (identical or nearly identical copies) or visual matches (images depicting similar scenes or entities). These results often provide contextual details such as publication dates, locations, and associated narratives, which help establish the provenance of the image. We also extract “about this image” metadata when available, which includes structured information such as the earliest known publication date or source. To reduce noise, we filter out unnecessary visual data like thumbnails while retaining essential textual context.
        \item \textbf{Claim-Based Web Search:} To capture the broader context surrounding the textual claim, we conduct Google searches using the claim text as a query. This allows us to gather a diverse set of news articles, reports, and scientific sources that explicitly discuss or reference the claim. Unlike reverse image search, which focuses on the image provenance, claim-based search emphasizes validating or contesting the factual statements made in the text. By integrating these results, we ensure that the reasoning process is informed by up-to-date and claim-specific evidence from reliable sources.
        \item \textbf{Fact-Checking Sites:} We incorporate evidence from established climate fact-checking organizations and curated databases. These sources provide direct assessments of claims, often labeling them as true, false, misleading, or lacking sufficient evidence. Fact-checking sites are particularly valuable because they offer expert-reviewed conclusions, supported by scientific consensus and references to primary data. Including this evidence strengthens the reliability of our system, as fact-checks serve as high-confidence signals in verifying or refuting claims.
        \item \textbf{GPT Searech:} This method is used to retrieve external evidence directly from the web in response to a textual claim. For each claim, the system issues a query and GPT generates a preview of the retrieved web results. Each preview consists of two parts: (1) a concise summary of the most relevant content, distilled from the retrieved page, and (2) the link to the original source. This approach allows us to efficiently capture key information without overwhelming the reasoning process with lengthy articles. Importantly, the linked source provides transparency and enables deeper verification when needed.
        Compared to full article retrieval, GPT Web Preview prioritizes coverage and efficiency. It ensures that every claim is associated with at least one relevant piece of external knowledge, even when fact-checks or provenance information are missing. By combining short, claim-centered summaries with direct source links, GPT Web Preview fills in evidence gaps and provides consistent background context across the dataset.
    \end{itemize}

The success rates for each external knowledge is shown in Figure~\ref{fig:retrieval_success_rate} for both dataset schemes.
\begin{figure}[h]
    \centering
    
    \begin{subfigure}{0.45\textwidth}
        \centering
        \includegraphics[width=\linewidth]{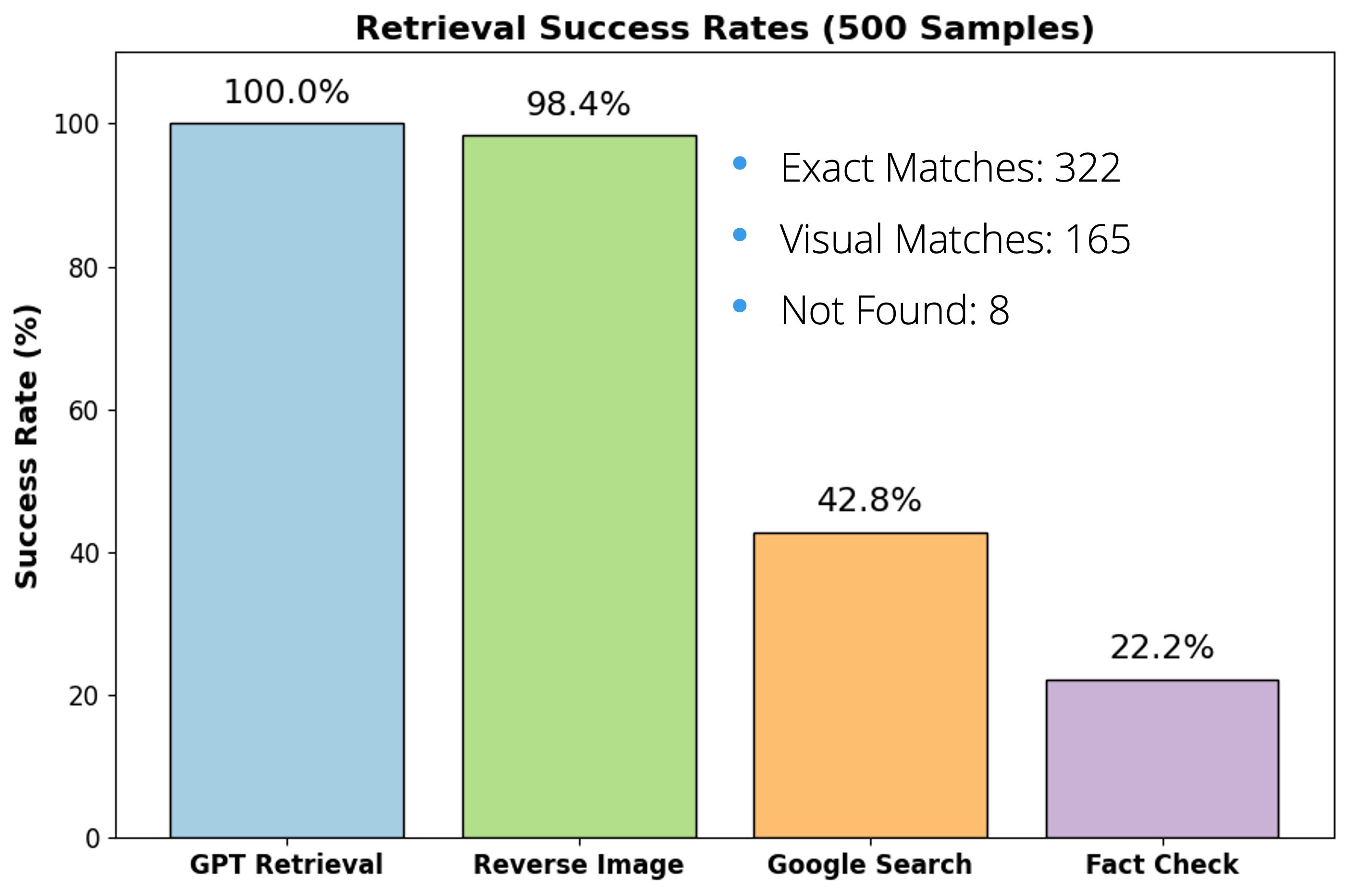}
        \caption{Across 4-class setup}
        \label{fig:4class_retrieval_success_rate}
    \end{subfigure}
    \begin{subfigure}{0.45\textwidth}
        \centering
        \includegraphics[width=\linewidth]{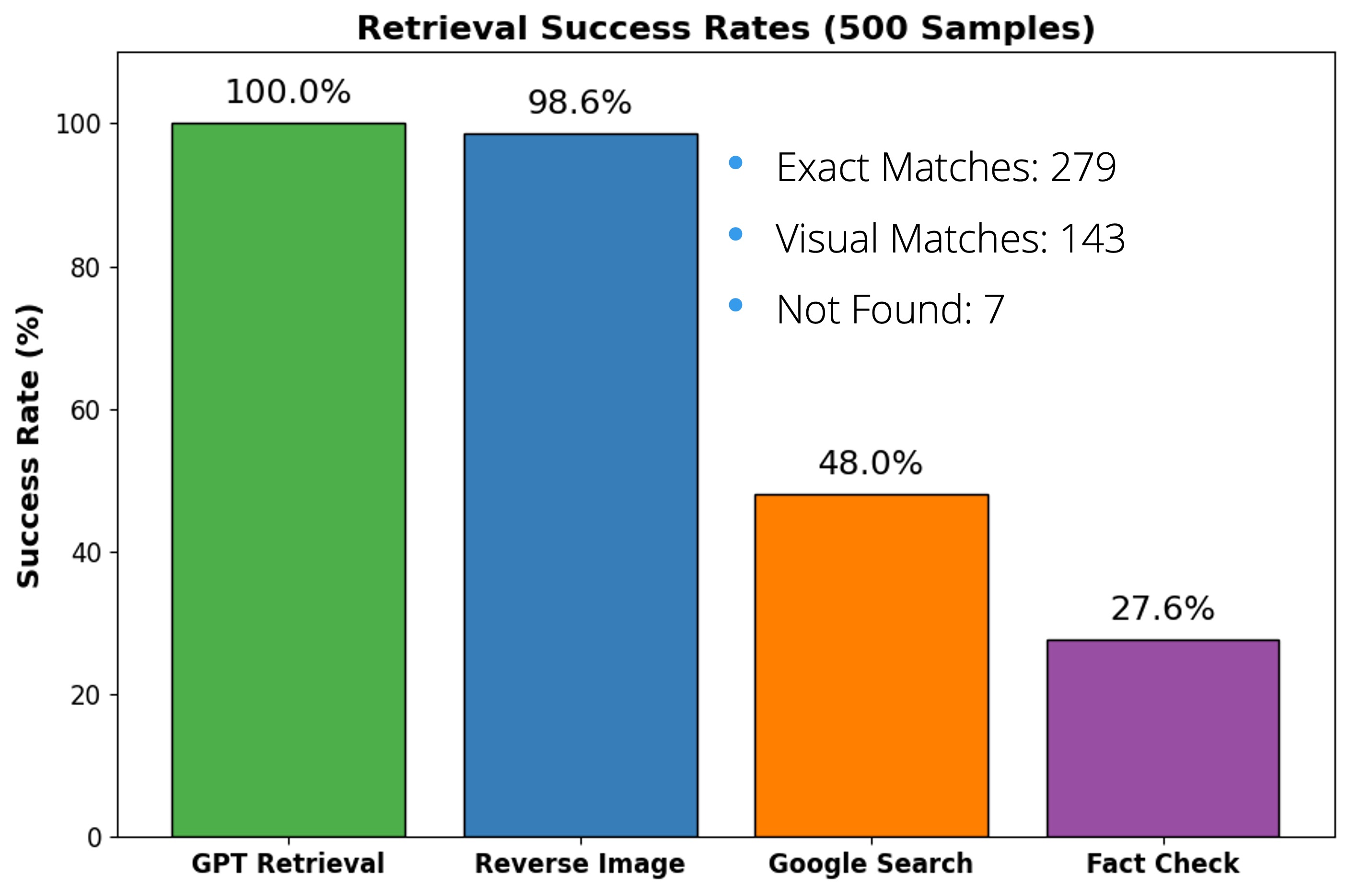}
        \caption{Across 2-class setup}
        \label{fig:2class_retrieval_success_rate}
    \end{subfigure}
    \caption{Retrieval Success Rate across}
    \label{fig:retrieval_success_rate}
\end{figure}

    \item \textbf{Reasoning Strategy:} We adopt both the Chain-of-Draft (CoD) and Chain-of-Thought (CoT) prompting strategies to compare the model’s reasoning depth, reliability, and efficiency in terms of time and token usage. 

    In the CoD approach, instead of directly providing a verdict, GPT-4o generates multiple reasoning drafts, each presenting a distinct interpretation of the image–text pair based on the available evidence. The model then critically evaluates these drafts to select the most coherent and factually supported explanation before assigning a final label. This iterative process encourages self-reflection, reduces impulsive judgments, and helps mitigate unsupported conclusions.
    
    In contrast, the CoT strategy relies on step-by-step reasoning where the model incrementally expands its thoughts to justify the final prediction. While CoT reasoning offers detailed logical progression, it typically consumes more tokens and time compared to CoD, as it explains each reasoning step explicitly rather than refining multiple drafts before finalizing the answer.

    \item \textbf{Label Assignment:} The model assigns one of four possible labels in the four-class setup: \textit{Accurate}, \textit{Misleading}, \textit{False}, or \textit{Unverifiable}, each accompanied by a justification. In the two-class setup, the labels are simplified to \textit{Accurate} and \textit{Disinformation}.

    \item \textbf{Source Comparison:} Each knowledge source (or combination of sources) is evaluated independently, enabling a comparison of model performance under different retrieval settings: internal-only, external-only, or combined. Our main methodology follows a conditional inclusion strategy, where external knowledge is incorporated in a prioritized order: fact-check sources (as the most reliable), followed by GPT search, reverse image search (with exact matches prioritized), and finally Google search. This conditional feeding of external knowledge prevents the model from being overwhelmed with excessive or less relevant information. Figure~\ref{fig:proposed_methodology} illustrates the overall methodology.
\end{itemize}

\begin{figure}[h]
    \centering
    \includegraphics[width=\linewidth]{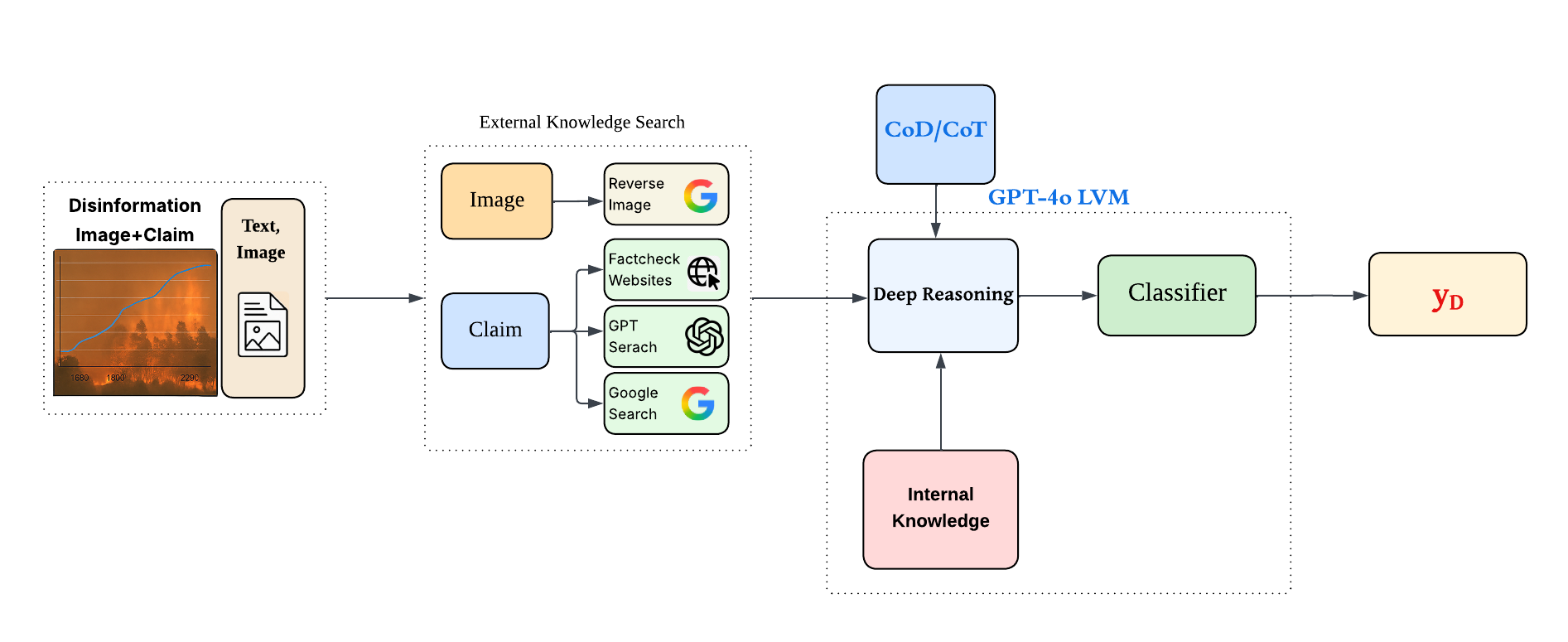}
    \caption{Conditional Inclusion Reasoning}
    \label{fig:proposed_methodology}
\end{figure}

\section{Experiments}
We evaluate GPT-4o’s ability to fact-check multimodal content using a two-stage protocol. 

In the first stage, the model is tested separately with each of the four external retrieval sources: (1) Google Reverse Image Search, (2) GPT-based web preview of the claim, (3) Google claim-based image search, and (4) trusted fact-checking websites. For each setting, GPT-4o receives the image, the associated claim, and the corresponding external evidence.  

In the second stage, all four external sources are combined into a unified context, and the model is prompted with the complete set of retrieved evidence. This setup allows us to examine whether integrating multiple sources enhances the accuracy and robustness of factual reasoning.  

Both stages are evaluated under the two reasoning strategies, Chain-of-Thought (CoT) and Chain-of-Draft (CoD).

\subsection{Evaluation Metrics}

We use both classification and behavioral metrics to assess the model's performance:

\begin{itemize}
    \item \textbf{Accuracy:} Percentage of predictions that exactly match the reference label.
    
    \item \textbf{Macro F1-Score:} Average F1-score computed across all four classes without weighting by class frequency.
    
    \item \textbf{Confusion Matrix:} Summarizes correct and incorrect predictions across all label categories.
    
    \item \textbf{Label Distribution:} Tracks the frequency of predicted labels to detect overuse of certain classes.
    
    \item \textbf{Rejection Rate:} Proportion of samples for which the model fails to return a valid label due to missing or inconclusive evidence.
    \[
    \text{Rejection Rate} = \frac{\# \text{Fallback Responses}}{\# \text{Total Samples}}
    \]
    \item \textbf{Confidence score:} We instructed the model to judge itself and assign how much confidance it is for its reasoning to have broader evaluation than only classification. this helps to evaluate how the reasoning is performed .
    
\end{itemize}

These metrics help quantify both factual correctness and the impact of different retrieval sources on GPT-4o’s reasoning.

\subsection{Results}
\begin{figure}[h]
    \centering
    \begin{subfigure}{0.43\textwidth}
        \centering
        \includegraphics[width=\linewidth]{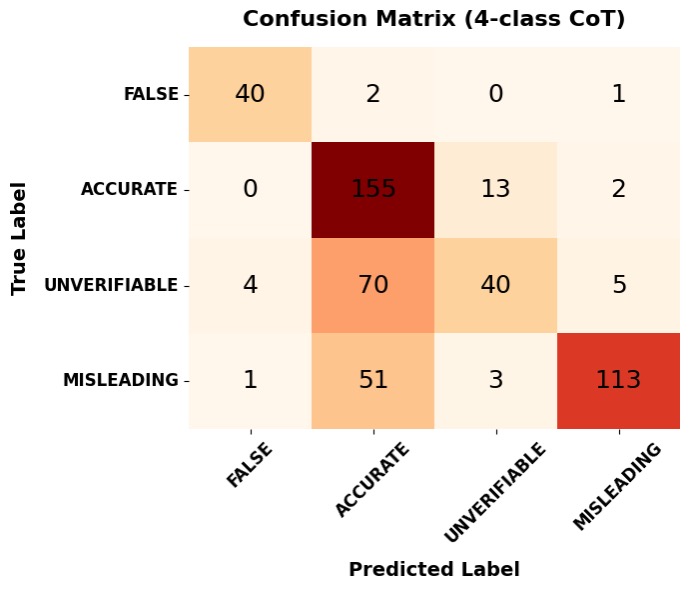}
        \label{fig:CoT in combination methodology}
    \end{subfigure}
    \begin{subfigure}{0.43\textwidth}
        \centering
        \includegraphics[width=\linewidth]{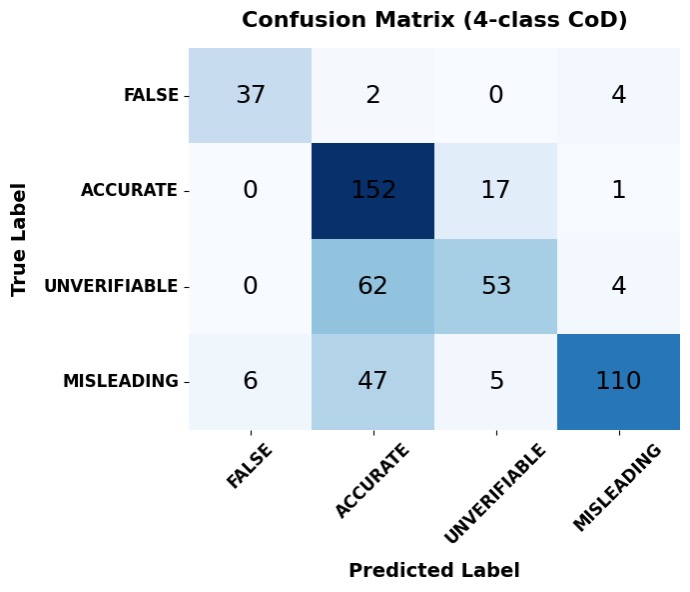}
        \label{fig:CoD in combination methodology}
    \end{subfigure}
    \caption{Confusion Matrix in Combination Source}
    \label{fig:confusion_matrix_4class}
\end{figure}

The comparisons between different sources of external knowledge based on different reasoning strategies using different metrics are presented in Tables ~\ref{tab:source_cot_metrics_4class}-~\ref{tab:source_cod_metrics_2class}. Confusion matrices are also provided for 4-class comparisons in Figure~\ref{fig:confusion_matrix_4class}. Token usage and running time details for different external sources and reasoning strategies are given in Table~\ref{tab:token_usage}.  
\begin{table}[h]
\captionsetup{skip=10pt} 
\centering
\renewcommand{\arraystretch}{1.3}
\begin{tabular}{|l|c|c|c|c|c|c|}
\hline
\rowcolor{yellow!20}
\textbf{Source CoT} & \textbf{Accuracy} & \textbf{Precision} & \textbf{Recall} & \textbf{F1} & \textbf{Rejection Rate} & \textbf{Confidence (avg)} \\
\hline
Fact check     & 66.60 & 69.88 & 68.16 & 68.18 & 1.2 & 79.56 \\
Google Search  & 60.60 & 65.92 & 60.87 & 62.29 & 2.0 & 80.46 \\
Reverse Image  & 62.00 & 67.89 & 60.99 & 62.32 & 0.8 & 86.11 \\
GPT Search     & 65.40 & 69.31 & 66.71 & 61.88 & 0.2 & 91.77 \\
Internal       & 63.80 & 77.30 & 69.06 & 68.35 & 2.6 & 77.75 \\
\textbf{Combination} & \textbf{69.60} & \textbf{77.37} & \textbf{71.27} & \textbf{71.01} & \textbf{0.0} & \textbf{88.72} \\
\hline
\end{tabular}
\caption{Performance metrics across different sources with CoT prompting in 4-class setup}
\label{tab:source_cot_metrics_4class}
\end{table}

\begin{table}[h]
\captionsetup{skip=10pt} 
\centering
\renewcommand{\arraystretch}{1.3} % wider rows
\begin{tabular}{|l|c|c|c|c|c|c|}
\hline
\rowcolor{blue!10}
\textbf{Source CoD} & \textbf{Accuracy} & \textbf{Precision} & \textbf{Recall} & \textbf{F1} & \textbf{Rejection Rate} & \textbf{Confidence (avg)} \\
\hline
Fact check     & 62.80 & 64.78 & 62.96 & 62.55 & 1.8 & 82.76 \\
Google Search  & 58.40 & 63.13 & 59.56 & 59.20 & 3.8 & 82.58 \\
Reverse Image  & 60.80 & 64.30 & 60.93 & 60.49 & 2.2 & 87.20 \\
GPT Search     & 69.00 & 74.24 & 71.55 & 66.86 & 0.6 & 91.68 \\
Internal       & 68.20 & 75.52 & 70.95 & 70.98 & 1.6 & 82.54 \\
\textbf{Combination} & \textbf{70.40} & \textbf{76.74} & \textbf{71.37} & \textbf{71.89} & \textbf{0.0} & \textbf{88.92} \\
\hline
\end{tabular}
\caption{Performance metrics across different sources with CoD prompting in 4-class setup}
\label{tab:source_CoD_metrics_4class}
\end{table}

\begin{table}[h]
\captionsetup{skip=10pt} 
\centering
\renewcommand{\arraystretch}{1.3}
\begin{tabular}{|l|c|c|c|c|c|c|}
\hline
\rowcolor{yellow!20}
\textbf{Source CoT} & \textbf{Accuracy} & \textbf{Precision} & \textbf{Recall} & \textbf{F1} & \textbf{Rejection Rate} & \textbf{Confidence (avg)} \\
\hline
Fact check     & 83.80 & 85.06 & 84.12 & 84.35 & 1.4 & 86.42 \\
Google Search  & 81.20 & 81.88 & 81.07 & 81.48 & 1.0 & 88.43 \\
Reverse Image  & 81.69 & 82.76 & 80.85 & 81.30 & 1.2 & 91.45 \\
GPT Search     & 84.00 & 84.63 & 83.46 & 83.86 & 0.4 & 93.38 \\
Internal       & 85.40 & 91.10 & 85.88 & 88.16 & 6.2 & 89.46 \\
\textbf{Combination} & \textbf{86.45} & \textbf{87.27} & \textbf{86.43} & \textbf{86.19} & \textbf{0.0} & \textbf{91.38} \\
\hline
\end{tabular}
\caption{Performance metrics across different sources with CoT prompting in 2-class setup}
\label{tab:source_cot_metrics_2class}
\end{table}

\begin{table}[h]
\captionsetup{skip=10pt} 
\centering
\renewcommand{\arraystretch}{1.3}
\begin{tabular}{|l|c|c|c|c|c|c|}
\hline
\rowcolor{blue!10}
\textbf{Source CoD} & \textbf{Accuracy} & \textbf{Precision} & \textbf{Recall} & \textbf{F1} & \textbf{Rejection Rate} & \textbf{Confidence (avg)} \\
\hline
Fact check     & 74.20 & 87.05 & 74.07 & 80.03 & 15.0 & 77.32 \\
Google Search  & 72.00 & 83.46 & 71.46 & 76.80 & 13.4 & 79.19 \\
Reverse Image  & 78.40 & 82.52 & 78.10 & 80.24 & 5.2  & 89.53 \\
GPT Search     & 83.80 & 85.36 & 83.08 & 83.84 & 1.0  & 92.59 \\
Internal       & 66.80 & 93.55 & 66.25 & 77.47 & 28.8 & 69.48 \\
\textbf{Combination} & \textbf{86.20} & \textbf{86.84} & \textbf{86.29} & \textbf{86.02} & \textbf{0.0} & \textbf{85.56} \\
\hline
\end{tabular}
\caption{Performance metrics across different sources with CoD prompting in 2-class setup}
\label{tab:source_cod_metrics_2class}
\end{table}

\begin{figure}[h]
    \centering
    \begin{subfigure}{0.35\textwidth}
        \centering
        \includegraphics[width=\linewidth]{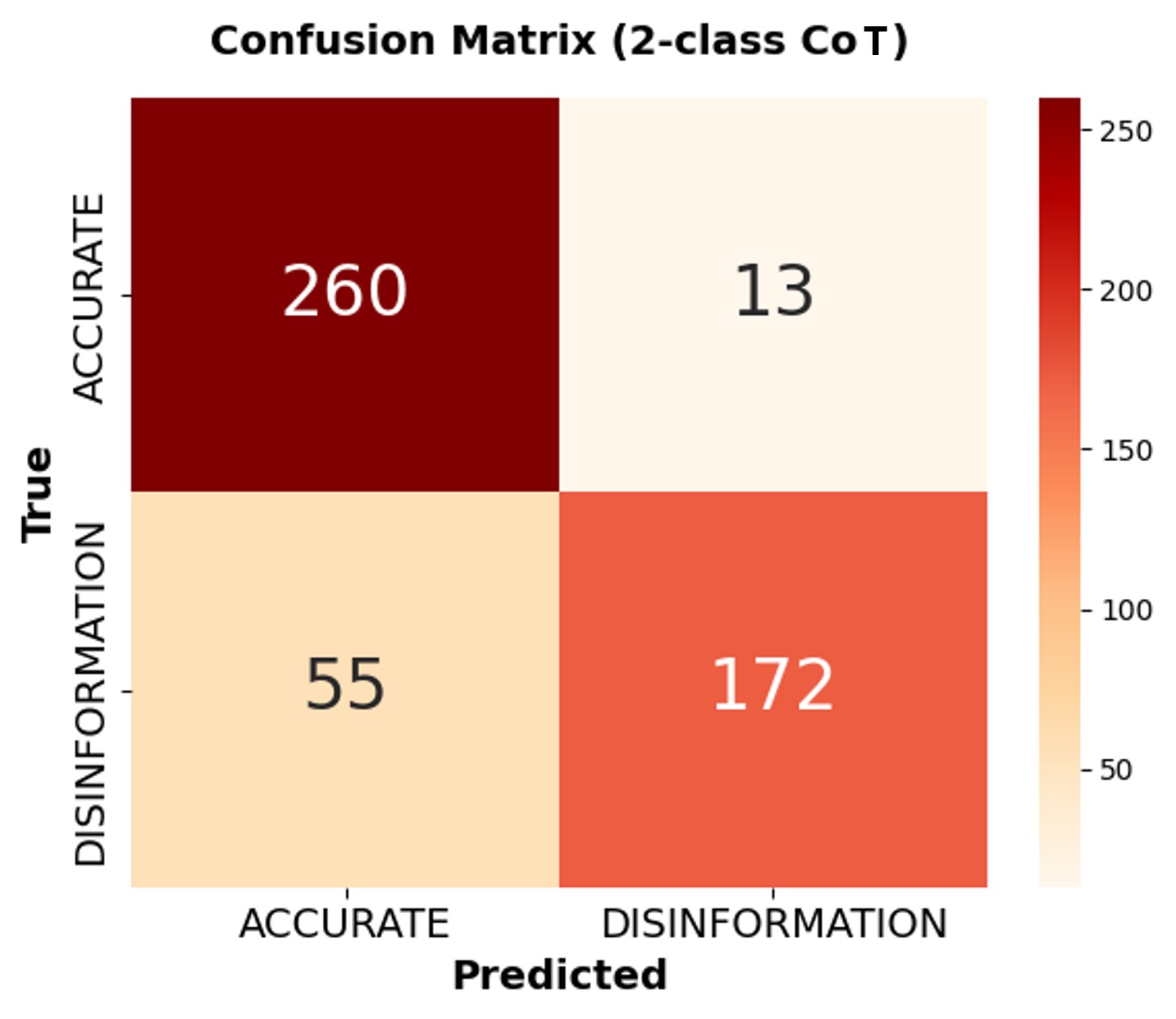}
        \label{fig:CoT in combination methodology}
    \end{subfigure}
    \begin{subfigure}{0.35\textwidth}
        \centering
        \includegraphics[width=\linewidth]{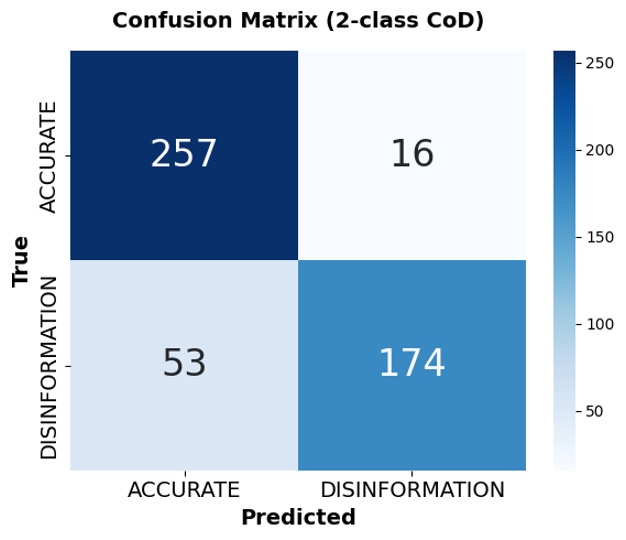}
        \label{fig:CoD in combination methodology}
    \end{subfigure}
    \caption{Confusion Matrix in Combination Source}
    \label{fig:confusion_matrix_2class}
\end{figure}

\begin{table}[h]
\captionsetup{skip=10pt} 
\centering
\renewcommand{\arraystretch}{1.3} % wider rows
\begin{tabular}{|l|c|c|c|c|}
\hline
\rowcolor{gray!20}
\textbf{Source} & \textbf{Strategy} & \textbf{Total Tokens (500 prompts)} & \textbf{Avg Prompt} & \textbf{Avg Time (s)} \\
\hline
GPT Search     & CoD & 800,007   & 1517.2 & 3.42 \\
               & CoT & 825,065   & 1538.2 & 4.09 \\
\hline
Reverse Image  & CoD & 1,451,923 & 2819.1 & 3.88 \\
               & CoT & 1,463,927 & 2840.1 & 3.81 \\
\hline
Factcheck Sites & CoD & 634,052   & 1184.6 & 3.67 \\
                & CoT & 652,833   & 1205.6 & 3.91 \\
\hline
Google Search  & CoD & 647,923   & 1212.1 & 3.67 \\
               & CoT & 665,816   & 1233.1 & 3.79 \\
\hline
Internal       & CoD & 583,505   & 1080.7 & 3.53 \\
               & CoT & 585,740   & 1088.7 & 3.65 \\
\hline
Combined       & CoD & $\sim$2,025,000 & 3964.5 & 5.10 \\
               & CoT & $\sim$2,048,050 & 4007.5 & 5.74 \\
\hline
\end{tabular}
\caption{Token usage, average prompt length, and average time across sources and strategies}
\label{tab:token_usage}
\end{table}

\subsection{Discussion}

These results indicate that when all four external sources are combined it outperforms the internal knowledge of the model obtained during training and also each individual external source. Furthermore, this combination led to zero rejection by model to fact check. The internal knowledge-based model, however, showed better results comparing to some of the individual external sources even though it has higher rejection rate. This is an indication that one individual external source may confuse the model and using them solo may not be enough with an exception of GPT Search. In terms of confidence, the model interestingly shows more confidence in its GPT Search than all sources together. We can also see that 4-class setup is more challenging for the model even using all external sources. Different reasoning strategies resulted similar performances with a little bit better results for CoD on 4-class dataset. In terms of cost, based on token usage, Reverse Image is the most expensive source while Factchecked Sites source is the cheapest in both CoT and CoD among individual external sources. When combined with other sources, it gets the heaviest overall with more than 2 million tokens used for 500 prompts/samples. Comparing different reasoning strategies, CoT generally uses more tokens than CoD.      

\section{Conclusion and futur work}
In this paper, we proposed to use different sources of external knowledge including reverse image search, factchecked sites, Google search and GPT search in order to improve the ability of vision-language
models VLMs to detect climate-related multimodal disinformation. Such external sources boost the performance of VLMs which rely only on their internal knowledge acquired during training. By retrieving up-to-date information, we increase their ability to reason about recent events or updates. In order to verify this approach, we automatically annotated 500 samples of the existing CLiME dataset, extracting disinformation related labels. This approach was then tested using GPT-4o on the new dataset, improving its performance using these external sources combined. Although, this enhancement in detecting climate disinformation is significant, we believe that there is still a lot of room for improvement. Future works are, hence, focused on annotating more data, optimizing the external sources especially on reverse image search and content-based image retrieval algorithms in order to feed VLMs with more informative images.    

\bibliographystyle{unsrt}
\bibliography{references}

@misc{coan2021,
  author       = {Coan, Travis G. and Boussalis, Constantine and Cook, John and Nanko, Mirjam O.},
  title        = {Computer-assisted detection and classification of misinformation about climate change},
  year         = {2021},
  month        = mar,
  day          = {9},
  url          = {https://doi.org/10.31235/osf.io/crxfm_v1},
  note         = {Preprint, OSF},
}

@article{meddeb2022,
  title={Counteracting French Fake News on Climate Change Using Language Models},
  author={Meddeb, Paul and Ruseti, Stefan and Dascalu, Mihai and Terian, Simina-Maria and Travadel, Sebastien},
  journal={Sustainability},
  volume={14},
  number={18},
  pages={11724},
  year={2022},
  publisher={MDPI},
  ISSN = {2071-1050},
  URL = {https://www.mdpi.com/2071-1050/14/18/11724},
  DOI = {10.3390/su141811724}
}

@misc{herasimenka2023,
      title={Promoting Reliable Knowledge about Climate Change: A Systematic Review of Effective Measures to Resist Manipulation on Social Media}, 
      author={Aliaksandr Herasimenka and Xianlingchen Wang and Ralph Schroeder},
      year={2024},
      eprint={2410.23814},
      archivePrefix={arXiv},
      primaryClass={cs.CY},
      url={https://arxiv.org/abs/2410.23814}, 
}

@misc{rojas2024,
      title={Augmented CARDS: A machine learning approach to identifying triggers of climate change misinformation on Twitter}, 
      author={Cristian Rojas and Frank Algra-Maschio and Mark Andrejevic and Travis Coan and John Cook and Yuan-Fang Li},
      year={2024},
      eprint={2404.15673},
      archivePrefix={arXiv},
      primaryClass={cs.LG},
      url={https://arxiv.org/abs/2404.15673}, 
}

@misc{fore2024,
      title={Unlearning Climate Misinformation in Large Language Models}, 
      author={Michael Fore and Simranjit Singh and Chaehong Lee and Amritanshu Pandey and Antonios Anastasopoulos and Dimitrios Stamoulis},
      year={2024},
      eprint={2405.19563},
      archivePrefix={arXiv},
      primaryClass={cs.CL},
      url={https://arxiv.org/abs/2405.19563}, 
}

@misc{allaham2025,
      title={Enhancing LLMs for Governance with Human Oversight: Evaluating and Aligning LLMs on Expert Classification of Climate Misinformation for Detecting False or Misleading Claims about Climate Change}, 
      author={Mowafak Allaham and Ayse D. Lokmanoglu and P. Sol Hart and Erik C. Nisbet},
      year={2025},
      eprint={2501.13802},
      archivePrefix={arXiv},
      primaryClass={cs.CY},
      url={https://arxiv.org/abs/2501.13802}, 
}

@misc{CliMe,
      title={CliME: Evaluating Multimodal Climate Discourse on Social Media and the Climate Alignment Quotient (CAQ)}, 
      author={Abhilekh Borah and Hasnat Md Abdullah and Kangda Wei and Ruihong Huang},
      year={2025},
      eprint={2504.03906},
      archivePrefix={arXiv},
      primaryClass={cs.CL},
      url={https://arxiv.org/abs/2504.03906}, 
}

@misc{lEMMA,
      title={LEMMA: Towards LVLM-Enhanced Multimodal Misinformation Detection with External Knowledge Augmentation}, 
      author={Keyang Xuan and Li Yi and Fan Yang and Ruochen Wu and Yi R. Fung and Heng Ji},
      year={2024},
      eprint={2402.11943},
      archivePrefix={arXiv},
      primaryClass={cs.CL},
      url={https://arxiv.org/abs/2402.11943}, 
}

@misc{CMIE,
      title={CMIE: Combining MLLM Insights with External Evidence for Explainable Out-of-Context Misinformation Detection}, 
      author={Fanxiao Li and Jiaying Wu and Canyuan He and Wei Zhou},
      year={2025},
      eprint={2505.23449},
      archivePrefix={arXiv},
      primaryClass={cs.MM},
      url={https://arxiv.org/abs/2505.23449}, 
}

@misc{shen2025gamed,
      title={GAMED: Knowledge Adaptive Multi-Experts Decoupling for Multimodal Fake News Detection}, 
      author={Lingzhi Shen and Yunfei Long and Xiaohao Cai and Imran Razzak and Guanming Chen and Kang Liu and Shoaib Jameel},
      year={2025},
      eprint={2412.12164},
      archivePrefix={arXiv},
      primaryClass={cs.LG},
      url={https://arxiv.org/abs/2412.12164}, 
}

@misc{wu2025,
      title={Seeing Through Deception: Uncovering Misleading Creator Intent in Multimodal News with Vision-Language Models}, 
      author={Jiaying Wu and Fanxiao Li and Min-Yen Kan and Bryan Hooi},
      year={2025},
      eprint={2505.15489},
      archivePrefix={arXiv},
      primaryClass={cs.CV},
      url={https://arxiv.org/abs/2505.15489}, 
}

\end{document}